\documentclass[journal]{IEEEtran}
\usepackage{amsmath,amsfonts}
\usepackage{algorithmic}
\usepackage{algorithm}
\usepackage{array}
\usepackage[caption=false,font=normalsize,labelfont=sf,textfont=sf]{subfig}
\usepackage{textcomp}
\usepackage{stfloats}
\usepackage{url}
\usepackage{verbatim}
\usepackage{graphicx}
\usepackage{cite}
\usepackage{lettrine}
\newcommand{\myparagraph}[1]{\vspace{0.1em}\noindent\textbf{#1}}
\usepackage[colorlinks,hyperindex,breaklinks]{hyperref}
\newcommand{\ie}{\textit{i}.\textit{e}.}
\newcommand{\eg}{\textit{e}.\textit{g}.}

\usepackage{pifont}
\newcommand{\cmark}{\ding{51}}%
\newcommand{\xmark}{\ding{55}}%
\usepackage{colortbl}
\usepackage{arydshln}
\usepackage{multirow}
\usepackage{multicol}
\usepackage{mathrsfs}
\usepackage{bm}
\usepackage{amsfonts}
\renewcommand{\baselinestretch}{0.98}
\begin{document}
\title{Synthetic Instance Segmentation from Semantic Image Segmentation Masks}
\author{
Yuchen Shen, 
Dong Zhang, Zhao Zhang, Liyong Fu, Qiaolin~Ye,~\IEEEmembership{Member,~IEEE} \vspace{-3em}
\thanks{This work was supported in part by the National Key Research and Development Program of China under Grant 2022YFD2201005 and in part by the National Science Foundation of China under Grants 62072246 and 32371877.}

\thanks{Y. C. Shen and Q. L. Ye, College of Information Science and Technology \& Artificial Intelligence, Nanjing Forestry University, Nanjing 210037, Jiangsu, China. E-mail: shenyuchen@njfu.edu.cn;yqlcom@njfu.edu.cn}

\thanks{D. Zhang, Department of Computer Science and Engineering, Hong Kong University of Science and Technology, Hong Kong, China. E-mail: dongz@ust.hk.}

\thanks{Z. Zhang, Key Laboratory of Knowledge Engineering with Big Data, Hefei University of Technology, Hefei 230009, China. E-mail: cszzhang@gmail.com.}

\thanks{L. Y. Fu, College of Information Science
and Technology, Nanjing Forestry University, Nanjing 210037, Jiangsu, China, and Institute of Forest Resource Information Techniques, Chinese Academy of Forestry, Beijing 100091, China. E-mail: fuliyong840909@163.com.}
}
\markboth{}
{Yuchen Shen \MakeLowercase{\textit{et al.}}: Synthetic Instance Segmentation from Semantic Image Segmentation Masks}
\maketitle
\begin{abstract}
In recent years, instance segmentation has garnered significant attention across various applications. However, training a fully-supervised instance segmentation model requires costly both instance-level and pixel-level annotations. In contrast, weakly-supervised instance segmentation methods, such as those using image-level class labels or point labels, often struggle to satisfy the accuracy and recall requirements of practical scenarios.
In this paper, we propose a novel paradigm
called Synthetic Instance Segmentation (SISeg). SISeg achieves instance segmentation results by leveraging image masks generated by existing semantic segmentation models, and it is highly efficient as we do not require additional training for semantic segmentation or the use of instance-level image annotations. In other words, the proposed model does not need extra manpower or higher computational expenses. Specifically, we first obtain a semantic segmentation mask of the input image via an existent semantic segmentation model. Then, we calculate a displacement field vector for each pixel based on the segmentation mask, which can indicate representations belonging to the same class but different instances, \ie, obtaining the instance-level object information. Finally, the instance segmentation results are refined by a learnable category-agnostic object boundary branch.
Extensive experimental results on two challenging datasets highlight the effectiveness of SISeg in achieving competitive results when compared to state-of-the-art methods, especially fully-supervised methods. 
The code will be released at: ~\href{https://github.com/ssyc123/SISeg}{SISeg}
\end{abstract}
\vspace{-2mm}\section{Introduction} 
\begin{figure}[t]
\centering
\includegraphics[scale=0.2]{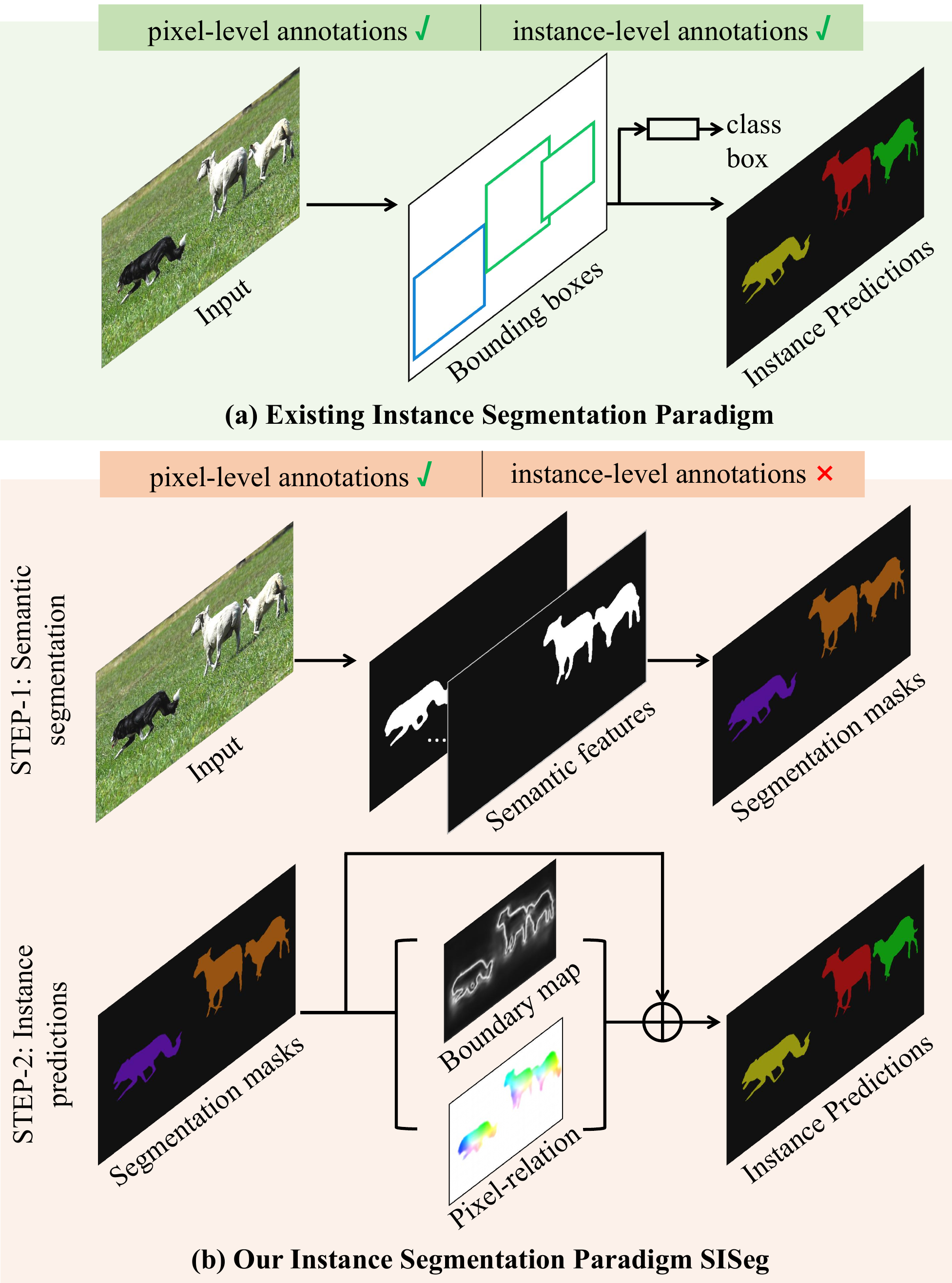}
\vspace{-2mm}
\caption{The diagrams illustrate the existing instance segmentation in (a) and our proposed approach SISeg in (b) of generating instance segmentation results from semantic image segmentation masks. Compared to the existing methods, ours does not require instance-level annotations and can be derived from existing semantic segmentation masks, resulting in higher efficiency. Samples are from the PASCAL VOC 2012 dataset~\cite{everingham2010pascal}.}
\vspace{-7mm}
\label{fig1}
\end{figure}
\IEEEPARstart{I}{nstance} segmentation is one of the fundamental and critical research tasks in communities of computer vision and multi-media, which aims at predicting pixel-level semantic classes and instance-level object masks~\cite{liu2018path, zhang2020feature, liu2021dance} of the given image. During the past years, instance segmentation has garnered significant attention, with several practical applications in a large range of scenarios, \eg, medical image analysis~\cite{zhang2022deep}, industrial robotics~\cite{schnieders2019fully}, and autonomous vehicles~\cite{yurtsever2020survey}. In particular, it is worth mentioning that the vision community has made remarkable progress in instance segmentation by leveraging both instance-level object detection and semantic segmentation tasks~\cite{peng2020deep, tian2020conditional, feng2023recurrent}. The implicit consensus behind this progress is that instance segmentation can be seen as a generalized
extension of object detection that involves results of instance-level object detection and pixel-level semantic segmentation, as illustrated in Figure~\ref{fig1} (a).
To this end, while top-down paradigms like Mask R-CNN~\cite{he2017mask}
can detect bounding boxes to localize individual instances and
assign semantic classes to the involved pixels within boxes,
they heavily rely on bounding-box proposal techniques (\eg, region proposal
networks~\cite{ren2015faster} and learnable reference points~\cite{carion2020end}) that may
not be suitable for some cases. For example, when the overlapping range of the
object is overmuch compact, unsatisfactory proposal boxes
may be obtained~\cite{kim2022beyond, zhang2020feature}. At the same time, bottom-up
paradigms like sequential grouping networks~\cite{liu2017sgn} achieve detailed segmentation results before separating object instances
using grouping post-processing~\cite{de2017semantic, kong2018recurrent, zhang2022e2ec}.
Although these paradigms naturally yield high-resolution masks, they fail to perform as well as the top-down methods mentioned above.

The currently dominant instance segmentation approaches have achieved impressive performance, but training an instance segmentation model is challenging, especially when compared to the elaborate semantic image segmentation model~\cite{zhang2020feature}. The presence of neighboring
or/and partially overlapping objects may further entangle this process~\cite{liu2018path, lin2017feature, jain2023oneformer}. Besides, the fully-supervised instance segmentation model requires high-quality manual annotations that are labor-intensive and time-consuming~\cite{pont2016multiscale, lin2014microsoft}. Such annotations involve not only collecting pixel-level annotations but also labeling instance-level annotations, \ie, labeling different instances of the same semantic class~\cite{li2022box2mask, tian2021boxinst}. As
shown in Figure~\ref{fig1}, for the given image containing many
animals, not only do we need to label objects of different
categories (\ie, the ``\textit{dog}'' and the ``\textit{sheep}''), but we also need
to label objects belonging to the same category but different
instances (\ie, the ``\textit{sheep1}'' and the ``\textit{sheep2}''). However, collecting annotations for instance segmentation tasks is extremely
expensive. For example, it is reported that annotation times for
pixel-level object masks are 79 seconds per instance for MS-COCO~\cite{lin2014microsoft} and about 1.5 hours per image for Cityscapes~\cite{cordts2016cityscapes}.
To alleviate this intractable problem, it is intuitively proposed
to train instance segmentation models using some easily
accessible annotations (\ie, the weakly-supervised instance
segmentation), such as image-level class annotations~\cite{hwang2021weakly, ahn2019weakly}
or point annotations~\cite{kim2022beyond, liao2023attentionshift}.

Nonetheless, there exists no free lunch when it comes to
weakly-supervised instance segmentation~\cite{zhu2019learning, lan2021discobox}. While
weakly-supervised annotations can offer approximate location
information of objects, the instance segmentation results may
have indistinct boundaries between objects, thereby affecting
the accuracy of downstream tasks. This is obvious, because
it is impossible for a machine model to automatically learn
information about different instances from the dataset~\cite{liao2023attentionshift, zhang2023weakly, kim2022beyond, ahn2019weakly}.
Even though some advanced feature extraction technologies
and post-processing methods have been proposed to mitigate
the limitations of the current experimental instance segmentation results, we contend that the weak supervision instance
segmentation setting alone cannot fundamentally address the
problem of inadequate instance segmentation precision.

In this paper, we propose a novel instance segmentation method called Synthetic Instance Segmentation (SISeg),
which can achieve satisfactory  results from the image masks, which are predicted by existing semantic segmentation models. Therefore, SISeg does not need to additionally train the model at all.
This capability also allows SISeg to eliminate the need for instance-level image annotations that are typically required by many existing instance segmentation methods, ultimately improving the efficiency of the model.
As illustrated in Figure~\ref{fig1} (b), in our implementation, \textbf{Step-1} first
uses the existing trained semantic segmentation model to obtain
a semantic segmentation mask of the input image, which only contains
pixel-level semantic information of different categories. Based on this, \textbf{Step-2} then achieves the purpose of
obtaining different instance information on this segmentation
mask by learning a displacement field vector for each pixel, which can indicate representations belonging to the same class but different instances. Finally, a learnable category-agnostic object
boundary branch is further deployed to refine the obtained
instance segmentation results. To validate
the effectiveness of our method, we deploy SISeg on two
fundamental yet challenging instance segmentation
datasets including PASCAL VOC 2012~\cite{everingham2010pascal} and ADE20K~\cite{zhou2017scene}.
Extensive experimental results demonstrate that
SISeg achieves competitive results compared to the state-of-the-art methods including fully-supervised instance segmentation techniques. Moreover, SISeg does not involve additional human resources or increased computational costs. The main contributions of this paper can be summarized as:
\begin{itemize}
\item[$\bullet$] We proposed a novel instance segmentation
method SISeg, which is based on the off-the-shelf semantic
image segmentation model that serves as the “Giant’s Shoulder”, and can be deployed on any semantic segmentation framework without changing the network architecture, which is a substantial change.
\item[$\bullet$] SISeg does not require instance-level image annotation or additional training for semantic segmentation by employing a displacement field detection module to provide instance-aware guidance to distinguish between object instances and a class boundary refinement module to detect distinct object boundaries to refine segmentation results.
\item[$\bullet$] Compared to the current state-of-the-art methods, which include fully-supervised instance segmentation models, SISeg achieves very competitive results in terms of accuracy and speed on two
challenging datasets including PASCAL VOC 2012 and
ADE20K with its efficient network structure.
\end{itemize}

The remainder of this paper is organized as follows. In Section~\ref{sec2}, some related works on instance segmentation, semantic segmentation, and weakly-supervised image segmentation methods are reviewed. Section~\ref{sec3} introduces our SISeg in details. Experiments results are reported in Section~\ref{sec4}, and conclusions are provided in Section~\ref{sec5}.

\vspace{-2.5mm}\section{Related works}
\label{sec2}
In this section, our aim is to review some related works on instance segmentation, semantic segmentation, and weakly-supervised image segmentation methods.
\vspace{-5mm}\subsection{Instance Segmentation}
The learning objective of instance segmentation is to assign a category label to each pixel of the input image while recognizing the instance information of objects involved~\cite{ yu2020sprnet, zhang2021semantic}. Currently, instance segmentation methods can be classified into the following two categories: {detection-based methods} and {segmentation-based methods}. \emph{\textbf{For the first category}}, the methods first detect the objective instance and then segment it by utilizing state-of-the-art object detection models~\cite{shen2023learning, wu2019dsn, zhu2020deformable, qiu2020borderdet} such as R-CNN, Fast/Faster R-CNN~\cite{ren2015faster}, where Mask R-CNN~\cite{he2017mask} and its follow-up work~\cite{liu2018path, huang2019mask, cheng2020boundary, tian2020conditional} are the representative methods. These methods add a mask prediction branch depending on the corresponding box results from Faster R-CNN and achieve objective segmentation performance on many benchmarks. 
However, the speed of these methods and the resolution of their masks are somewhat limited due to the time-consuming feature re-pooling operation that inevitably loses spatial information. 
\emph{\textbf{For the second category}}, the methods first predict pixel-wise masks and then group instance-level objects together based on a set of instance cues~\cite{peng2020deep, zhang2022e2ec}, \eg, pixel-pair relations~\cite{kong2018recurrent} and instance boundaries~\cite{yin2021bridging}. In some advanced methods~\cite{de2017semantic, liu2017sgn},
instance masks are formed by learning the pixel association
embedding, \ie, the embedding of the pixels belonging to the same instance is similar and advanced. Our approach also falls in the second category. Different from the above methods trained with instance-level and pixel-level supervisions, our approach uses pixel-level annotations as the unique supervision, which bridges the gap between semantic segmentation and instance segmentation while maintaining a better accuracy-annotations trade-off.
\vspace{-5mm}\subsection{Semantic Segmentation}
Semantic segmentation can be seen as a special case of instance segmentation, with a category label assigned to each pixel~\cite{zhang2021self}.
FCN~\cite{long2015fully} is a pioneering method that adopts fully convolutional neural network for semantic segmentation.
After that, FCN-based methods have been extensively proposed~\cite{zhang2023augmented}, which can be divided into those that enlarge the receptive field~\cite{chen2017rethinking,zhao2017pyramid}, and those that use construct diverse attention modules~\cite{cheng2022masked,yuan2020object}.
Concretely, to maintain a high feature map resolution, models (\eg, DeepLabv3~\cite{chen2017rethinking} and PSPNet~\cite{zhao2017pyramid}) enlarge the receptive field by introducing dilated convolution and spatial pyramid pooling, such that the output feature maps have more global and semantic information. 
More methods show promising results on this task at the expense of constructing
many various attention modules~\cite{cheng2022masked,yuan2020object}, making the
resulting framework low efficiency.  
Besides, SETR~\cite{zheng2021rethinking} and its derivative SegFormer~\cite{xie2021segformer} have been developed based on Vision Transformer (ViT)~\cite{dosovitskiy2020image}.
Our proposed SISeg is flexible and efficient, making it applicable to off-the-shelf semantic segmentation models without additional modifications.  
\vspace{-4mm}\subsection{Weakly-Supervised Image Segmentation}
Since it is expensive to obtain large-scale instance-level and pixel-level annotations, recent effort is devoted to adopting a weak supervisions, such as image-level class labels~\cite{ahn2019weakly,zhu2019learning, pont2016multiscale}, bounding boxes~\cite{lan2021discobox, tian2021boxinst} or points~\cite{liao2023attentionshift,kim2022beyond}.
\emph{{For image-level labels}}, most of the initial work for weakly-supervised instance segmentation has been developed on class activation maps~\cite{zhu2019learning} or on segmentation proposals~\cite{pont2016multiscale}.
\emph{{For bounding boxes}}, 
based on CondInst~\cite{tian2020conditional}, BoxInst~\cite{tian2021boxinst} is proposed to exploit pixel similarity with a projection loss.
DiscoBox~\cite{lan2021discobox} further leverages pair-wise relation loss to generate proxy mask labels.
\emph{{For points}}, proposal-based methods~\cite{kim2022beyond, liao2023attentionshift} were presented with single or multiple points per instance as supervision.
Although the existing weakly-supervised image segmentation methods can achieve certain results, they are still insufficient when facing requirements for high accuracy in realistic scenarios. In this paper, we propose to generate instance segmentation results from existing trained semantic segmentation models.
Compared to fully-supervised instance segmentation models, our approach can also be viewed as a weakly supervised setting, \ie, our approach only requires pixel-level image annotations.

\vspace{-2mm}\section{Our Approach}
\label{sec3}
In this section, we specifically describe the proposed SISeg.
Firstly, 
we introduce the preliminaries of instance segmentation. Then, the overall framework of our presented SISeg model is shown. 
Afterwards, we elaborate in detail about our proposed displacement field detection module, class boundary refinement module and semantic-aware propagation module, respectively. 
Finally, the overall loss function and discussions on the comparisons with the most related models  are described.
\vspace{-4mm}\subsection{Preliminaries}
\label{sec3:1}
Given a training image $\boldsymbol{X} \in \mathbb{R}^{H \times W \times 3}$ with both pixel-level annotation \( \boldsymbol{\hat{X}_P} \in \mathbb{R}^{H \times W \times 1} \) and instance-level annotation \( \boldsymbol{\hat{X}_I} \in \mathbb{R}^{H \times W \times 1} \) (\ie, ${H \times W}$ denotes the spatial size, and $3$ or $1$ is the number of channels), the objective of instance segmentation model is to predict an instance-level object mask $\boldsymbol{M}_{ins} \in \mathbb{R}^{H \times W \times 1}$ with pixel-level semantic classes. 
To this end, some existing work breaks up the task of instance segmentation into two sub-tasks so that its combinatorial complexity can be avoided. 
The input image is first taken into a semantic segmentation model to output a semantic mask $\boldsymbol{M}_{sem} \in \mathbb{R}^{H \times W \times 1}$, which is then fed into the post-processing procedure (\eg, metric learning, clustering) to predict the corresponding instance segmentation.
Our SISeg also follows this stream. 

Although instance segmentation approaches have achieved remarkable performance over recent years, these fully-supervised methods often have the problem that acquiring instance-level manual annotations is extremely tedious and time-consuming. To alleviate this problem, lots of effort has been made on weakly-supervised instance segmentation, \ie, using easily accessible annotations. 
On the other hand, these weakly-supervised models still suffer from inferior segmentation accuracy because of coarse-grained annotations that fail to provide specific location information of objects.
Therefore, we propose an effective and efficient approach, termed SISeg, which generates high-quality instance segmentation results from off-the-shelf semantic segmentation models and achieves a better accuracy-annotations trade-off \textbf{\emph{by using pixel-level annotations only}}. 

\begin{figure*}[t]
\centering
\includegraphics[width=1\textwidth]{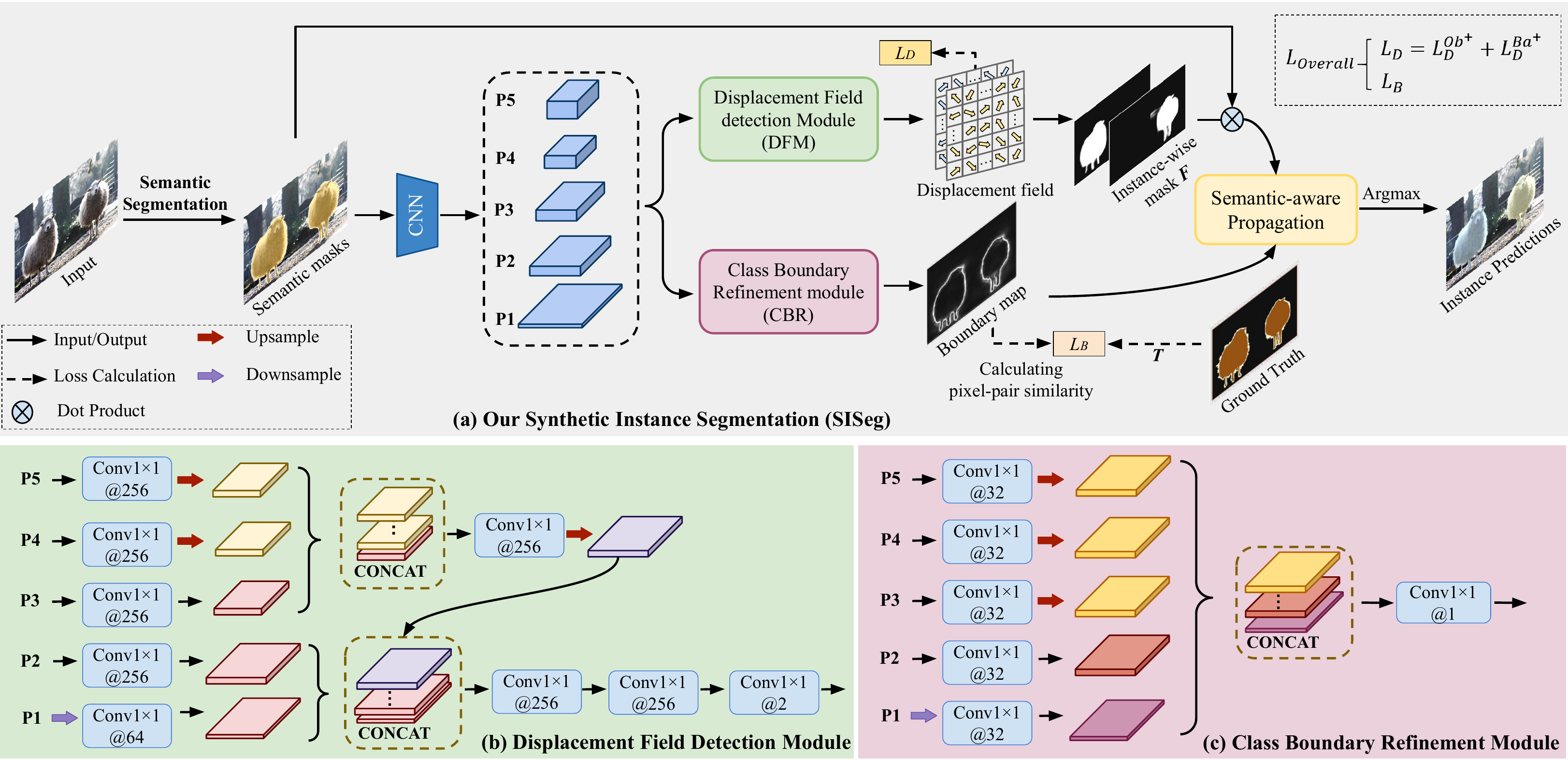}
\vspace{-6mm}
\caption{
The architecture of our proposed SISeg (a) mainly consists of two parallel branches sharing a backbone: the DFM branch (b) for separating the instances via predicting the displacement field and the CBR branch (c) for improving segmentation results along the boundary via calculating the semantic similarity of pixel pairs, which collaboratively achieve corresponding instance segmentation from the existing semantic masks without instance-level annotations. 
The whole network is optimized
by minimizing both losses $L_D$ and $L_B$ on pixel-level annotated data. The $T$ function is used to transform the ground truth label map into semantic similarity labels for calculating the $L_B$.
The samples are selected from the \emph{val} set of PASCAL VOC 2012~\cite{everingham2010pascal}. 
}

\vspace{-5mm}
\label{fig2}
\end{figure*}

\vspace{-4mm}\subsection{Overview}
\label{sec3:2}
As shown in Figure~\ref{fig2} (a), our proposed SISeg is 
a two-step framework that
mainly has four components: a well-trained semantic segmentation model, a Displacement Field detection Module (DFM), a Class Boundary Refinement module (CBR), and a semantic-aware propagation module.
In the beginning, for an input image $\boldsymbol{X} \in \mathbb{R}^{H \times W \times 3}$ 
with the pixel-level label \( \boldsymbol{\hat{X}_P} \in \mathbb{R}^{H \times W \times 1} \), 
the common semantic mask is derived from a well-trained semantic segmentation model. 
After that, we send the obtained semantic segmentation into our proposed SISeg, which has two parallel branches. 
The DFM branch is responsible for making a distinction between the different instances belonging to the same
class by predicting a displacement field vector for each pixel. 
Next, the instance-level object mask can be acquired by 
re-assigning class labels based on the semantic mask.
Another CBR branch aims to explore explicit boundary information by learning the semantic similarity between a pair of pixels, which are propagated with random walk to the entire semantically identical region of the obtained instance-level mask.
Finally, edge-enhanced instance segmentation is
generated by selecting the combination of class and instance labels associated with each highest-scoring pixel.

\vspace{-4mm}\subsection{Synthetic Instance Segmentation (SISeg)}
\label{sec3:3}

\subsubsection{Displacement Field Detection}
\label{sec3:3：1}
\textbf{Network architecture.}
In order to capture sufficient instance-aware cues, as illustrated in Figure~\ref{fig2} (b), we propose a DFM.
For a semantic segmentation mask $\boldsymbol{M}_{sem} \in \mathbb{R}^{H \times W \times 1}$ as input, 
we start by exploiting a CNN (\eg, ResNet-50/101~\cite{he2016deep}) to 
extract a set of feature maps.
The output of the CNN is then used to construct the feature pyramid $\left\{P_1, P_2, P_3, P_4, P_5\right\}$.
We apply a \( 1 \times 1 \) convolution layer to each feature map individually, reducing it to 256 if the channel dimension is larger than that. Similar to~\cite{lin2017feature}, we integrate multi-level feature maps in a top-down manner to improve the prediction of the displacement fields.
Concretely, 
we upsample the feature maps from the top two levels to keep their resolution 
aligned with that of 
$P_3$, which are then concatenated and passed through 256 filters of size \( 1 \times 1 \) followed by the upsampling operation to align with $P_2$. 
We formulate this procedure as: 
\vspace{-2mm}\begin{equation}
P_{\mathrm{cat}}^{\mathrm{D}}= \mathrm{E}\left(\operatorname{Cat}\left[\mathrm{E}\left(P_3\right) ; \mathrm{E}\left(P_4\right)\uparrow^2 ; \mathrm{E}\left(P_5\right)\uparrow^2\right]\right)\uparrow^2
\end{equation}
where $\mathrm{E}(\cdot)$ represents feature embedding operation (1 $\times$ 1 convolution + GroupNorm + ReLu~\cite{glorot2011deep}).
$\operatorname{Cat}[\cdot;\cdot]$ means channel-wise concatenation. 
$\uparrow^s$ denotes the upsampling operation with scale factor $s$.
For the feature map from the bottom level \( P_{1} \), we downsample its spatial resolution by a factor of 2.
After that, 
we concatenate the feature maps of three branches to
acquire the multi-scale feature maps.
Finally, a series of 1 $\times$ 1 convolutions are appended to predict an displacement field $\mathcal{D} \in \mathbb{R}^{\frac{H}{4} \times \frac{W}{4} \times 2}$, 
where the learned 2D offset vector for each pixel points to its corresponding instance centroid. It can be given by:
\vspace{-1mm}
\begin{equation}
\mathcal{D}=f^{1 \times 1}\left(\mathrm{E}\left(\mathrm{E}\left(\operatorname{Cat}\left[\mathrm{E}\left(P_{1}\downarrow_2\right); 
\mathrm{E}\left(P_{2}\right)
; P_{\mathrm{cat}}^{\mathrm{D}}\right]\right)\right)\right),
\end{equation}
where $f^{1 \times 1}$ is a standard 1 $\times$ 1 convolutional layer. $\downarrow_s$ denotes the downsampling operation with scale factor $s$.
In this way, the gap between semantic information and structural details can be bridged, and our DFM could access rich instance-sensitive information that is then used for subsequent instance centroid detection as well as the generation of binary segmentation masks for each instance.
\

\textbf{Learning.}
Given a semantic mask $\boldsymbol{M}_{sem} \in \mathbb{R}^{H \times W \times 1}$ with its pixel-level labels \( \boldsymbol{\hat{X}_P} \in \mathbb{R}^{H \times W \times 1} \), 
we first sample pixels that lie within a specified distance $\theta$ in pairs: 
\vspace{-1mm}
\begin{equation}
\mathcal{S}=\left\{(\alpha, \beta) \mid\left\|\mathbf{v}_{\alpha}-\mathbf{v}_{\beta}\right\| - \theta <0, \forall \alpha \neq \beta \right\},
\end{equation} 
where $(\alpha, \beta)$ is a pair of pixels,
$\mathbf{V}=\left\{\mathbf{v}_{\alpha}, \mathbf{v}_{\beta}, \ldots, \mathbf{v}_{i}, \ldots,\mathbf{v}_{n}\right\}$ denotes a set of pixel coordinate vectors, 
and $\|\cdot\|$ is the $\ell_2$ distance.
On the basis of $\boldsymbol{M}_{sem}$, 
DFM prefers predictions that are sensitive to different object instances of the same class. 
It means that regions representing different classes determined by $\boldsymbol{M}_{sem}$ can be ignored. 
Therefore, we further divide $\mathcal{S}$ into three subsets according to the class equivalence:
$\mathcal{S}_{\mathrm{Ob}}^{+}$ for objects, $\mathcal{S}_{\mathrm{Ba}}^{+}$ for background, and
$\mathcal{S}^{-}$.
The pixel pair is classified into $\mathcal{S}_{\mathrm{Ob}}^{+}$ if their class IDs $C$ are the same and non-zero.
If the same as zero, they are classified into $\mathcal{S}_{\mathrm{Ba}}^{+}$, and $\mathcal{S}^{-}$ otherwise:
\vspace{-3mm}
\begin{equation}
\mathcal{S}_{\mathrm{Ob}}^{+}=\left\{(\alpha, \beta) \mid C_{\alpha}=C_{\beta} \neq 0,(\alpha, \beta) \in \mathcal{S}\right\},
\end{equation} 
\vspace{-4mm}
\begin{equation}
\mathcal{S}_{\mathrm{Ba}}^{+}=\left\{(\alpha, \beta) \mid C_{\alpha}=C_{\beta} = 0,(\alpha, \beta) \in \mathcal{S}\right\},
\end{equation} 
\vspace{-4mm}
\begin{equation}
\mathcal{S}^{-}=\left\{(\alpha, \beta) \mid C_{\alpha} \neq C_{\beta},(\alpha, \beta) \in \mathcal{S}\right\}.
\end{equation} 
For the predicted displacement field $\mathcal{D} \in \mathbb{R}^{\frac{H}{4} \times \frac{W}{4} \times 2}$, 
all pixels for objects are associated with their corresponding instances by iteratively learning offset vectors, \ie, $I_j=\mathbf{v}_i+\mathcal{D}({\mathbf{v}_i})$.
Typically, the $j$-th instance centroid $\hat{I_j}$ can be extracted by 
$
\hat{I_j}=\frac{1}{n} \sum_{\mathbf{v} \in K_j} \mathbf{v},
$
where $K=\left\{K_1, K_2, \ldots, K_N\right\}$ denotes a set of instances.
Also, it is bound by the loss function with direct supervision:
\vspace{-2mm}\begin{equation}\label{eq7}
\mathcal{L}_{\mathcal{D}}^\mathrm{dire}=\sum_{i=1}^n\left\|\mathcal{D}({\mathbf{v}_i})-(\hat{I_j}-\mathbf{v}_{i})\right\|, 
\quad \mathbf{v}_{i} \in K_j.
\vspace{-1mm}\end{equation} 
It is important to acknowledge that, despite the absence of direct instance-level supervision, we still argue that the 2D offset vector can be learned via a self-supervised manner. This is designed based on the key assumption that a pair of pixels belonging to the same object will exhibit 2D offset vectors 
that converge towards the identical instance centroid.
Concretely, for every foreground pixel pair $(\alpha, \beta) \in \mathcal{S}_{\mathrm{Ob}}^{+}$ that are sampled in a small range, we believe that they are part of the same instance, 
which is constrained by the definition of centroid. Consequently, we formulate the learning process for each instance as follows.
\vspace{-1mm}\begin{equation}
\begin{split}
&\mathbf{v}_{\alpha}+\mathcal{D}({\mathbf{v}_{\alpha}})=\mathbf{v}_{\beta}+\mathcal{D}({\mathbf{v}_{\beta}}),\\
&\text{s.t.} \sum_{\mathcal{D} \in K_j} \mathcal{D}=0, \quad\forall j \in\{1, 2, \ldots, N\},
\vspace{-1mm}\end{split}
\end{equation} 
where $\mathbf{v}_{\alpha}+\mathcal{D}({\mathbf{v}_{\alpha}})$ means
shifting the pixel position $\mathbf{v}_{\alpha}$ by the learned 2D offset vector $\mathcal{D}({\mathbf{v}_{\alpha}})$.
This means that the ideal condition is
$
\mathbf{v}_{\alpha}-\mathbf{v}_{\beta}=\mathcal{D}({\mathbf{v}_{\beta}})-\mathcal{D}({\mathbf{v}_{\alpha}}).
$ 
Motivated by Eq.~\ref{eq7}, we use $L_1$ loss to minimize the discrepancy between $\hat{\mathcal{D}}_{\alpha, \beta}$ and $\mathcal{D}_{\alpha, \beta}$, which is activated at positive pixel pairs belonging to object instances:
\vspace{-1mm}\begin{equation}
\mathcal{L}_{\mathcal{D}}^\mathrm{{{Ob}^{+}}}=\frac{1}{\left|\mathcal{S}_{\mathrm{Ob}}^{+}\right|} \sum_{(\alpha, \beta) \in \mathcal{S}_{\mathrm{Ob}}^{+}}|\mathcal{D}_{\alpha, \beta}-\hat{\mathcal{D}}_{\alpha, \beta}|,
\vspace{-2mm}\end{equation} 
where $\hat{\mathcal{D}}_{\alpha, \beta} = \mathbf{v}_{\alpha}-\mathbf{v}_{\beta}$ represents the difference in pixel coordinates, and $\mathcal{D}_{\alpha, \beta} = \mathcal{D}({\mathbf{v}_{\beta}})-\mathcal{D}({\mathbf{v}_{\alpha}})$ is their vector displacement.
As for the positive pairs from the background $(\alpha, \beta) \in \mathcal{S}_{\mathrm{Ba}}^{+}$,
our goal is to suppress the estimation of their scattered centroids, which would mess up ones extracted from the foreground. The loss function $\mathcal{L}_{\mathcal{D}}^\mathrm{{{Ba}^{+}}}$ is defined as:
\vspace{-2mm}\begin{equation}
\mathcal{L}_{\mathcal{D}}^\mathrm{{{Ba}^{+}}}=\frac{1}{\left|\mathcal{S}_{\mathrm{Ba}}^{+}\right|} \sum_{(\alpha, \beta) \in \mathcal{S}_{\mathrm{Ba}}^{+}}|\mathcal{D}_{\alpha, \beta}|.
\end{equation}
However, supervised by weak labels, the initial displacement field $\mathcal{D}^0$ trained by DFM is inaccurate. To this end, we 
take an iterative approach to refine $\mathcal{D}$ following the form:
\vspace{-1mm}\begin{equation}\label{eq11}
\mathcal{D}^{U+1}(\mathbf{v})=\mathcal{D}^U(\mathbf{v})+\mathcal{D}{(\mathbf{v}+\mathcal{D}^{U}(\mathbf{v}))}, \quad \forall \mathbf{v},
\end{equation}
where $U \in\{0, 1, \ldots, u\}$ is an iteration index. It iteratively refines each displacement vector based on the previously estimated displacement vectors.
Then, the centroid of each instance is extracted by clustering 2D offset vectors from the refined displacement field. Actually, we regard a small set of neighboring pixels whose displacement vectors are close to zero in magnitude 
as candidate centroids, followed by the connected component labeling (CCL)~\cite{he2009fast} algorithm to extract the centroid of each instance. Last, the binary mask for each instance
$\boldsymbol{F} \in \mathbb{R}^{\frac{H}{4} \times \frac{W}{4} \times N}$ (\ie, \( N \) means the number of instances) can be attained by: 
\vspace{-2mm}\begin{equation}
\boldsymbol{F}^{(j)}(\mathbf{v})= \begin{cases} 1 & \text { if } \mathbf{v}+\mathcal{D}(\mathbf{v}) \in K_j, \quad \forall \mathbf{v} \\ 0 & \text { otherwise }\end{cases}
\end{equation}
where $j \in\{1, 2, \ldots, N\}$ is the instance index.
As there is no category information in the binary instance-wise mask $\boldsymbol{F}$, 
we associate it with the input semantic mask $\boldsymbol{M}_{sem}$ that provides the corresponding semantic class labels. 
The class-specific instance mask 
$\bar{\boldsymbol{F}} \in \mathbb{R}^{C \times N \times \frac{H}{4} \times \frac{W}{4}}$
can be given by: 
\vspace{-2mm}\begin{equation}
\bar{\boldsymbol{F}}(\textit{} { Score }) =
\boldsymbol{F}(\textit{} { Objectness }) \times \boldsymbol{M}_{sem}(\textit{} { Class }),
\vspace{-2mm}\end{equation}
where $\boldsymbol{F}(\textit{} { Objectness })$ is objectness score from the instance-wise mask $\boldsymbol{F}$, and $\boldsymbol{M}_{sem}(\textit{} { Class })$ is normalized classification score.   

\subsubsection{Class Boundary Refinement}
\label{sec3:3：2}
\textbf{Network architecture.}
Semantic segmentation, as the first step of our network, is important for the subsequent identification of instances. 
However, its performance is not always desirable. 
For this purpose, we design a CBR module, as depicted in Figure~\ref{fig2} (c), which generates sharp boundaries between classes as supplementary information by leveraging the only semantic label maps $\boldsymbol{\hat{X}_P}$ available.
Likewise, with $\boldsymbol{M}_{sem}$ as input,
CBR shares the feature maps $\left\{P_1, P_2, P_3, P_4, P_5\right\}$ from the backbone with DFM to make the whole network more efficient.
More specifically, 
1 × 1 convolutions are utilized followed by a simple bilinear upsampling by a factor of 2 or 4.
Then, we concatenate these feature maps derived from different levels and fetch the boundary map $\mathcal{B} \in \mathbb{R}^{H \times W \times 1}$ through a  1 × 1 convolution. That is,
\vspace{-2mm}\begin{equation}
\begin{aligned}
\mathcal{B}=f^{1 \times 1}(\operatorname{Cat}[ & \mathrm{E}\left(P_1 \downarrow_2\right) ; \mathrm{E}\left(P_2\right) ; \\
& \left.\left.\mathrm{E}\left(P_3\right) \uparrow^2 ; \mathrm{E}\left(P_4\right) \uparrow^4 ; \mathrm{E}\left(P_5\right) \uparrow^4\right]\right).
\vspace{-1.5mm}\end{aligned}
\end{equation}\vspace{-3mm}
\

\textbf{Learning.}
On top of the generated boundary map $\mathcal{B}$, whose values are limited between $[0, 1]$ (\ie, 1 represents the boundary), we further measure the similarity within pairs $(\alpha, \beta)$ using the following formula:
\vspace{-2mm}\begin{equation}
r_{\alpha \beta}=(1-\max _{i \in \left\{\gamma, \ldots, \eta\right\}} \mathcal{B}\left(\mathbf{v}_i\right))^\lambda,
\vspace{-1.5mm}\end{equation} 
where $\left\{\gamma, \ldots, \eta\right\}$ is a collection of pixels on the line from $\mathbf{v}_\alpha$ to $\mathbf{v}_\beta$, and $\lambda > 1$ that ignores trivial values.
Besides, we define $\mathcal{T}$ function to transform ground truth similarity label map $\mathcal{\hat{B}}$ from $\boldsymbol{\hat{X}_P}$,
which assigns binary labels to the sampled pairs. More concretely, their similarity label is 1 if the class id of a given pair $C_{\alpha}, C_{\beta}$ are the same, and 0 otherwise. 
Then, the class-balanced cross-entropy loss is used as the objective function for training similarity:
\vspace{-2mm}\begin{equation}
\begin{aligned}
\mathcal{L}_{\mathcal{B}} & =   \mathcal{L}_{\mathcal{B}}^\mathrm{{{Ob}^{+}}} + \mathcal{L}_{\mathcal{B}}^\mathrm{{{Ba}^{+}}} +
2\mathcal{L}_{\mathcal{B}}^\mathrm{-}
\\& =  
-\sum_{(\alpha, \beta) \in \mathcal{S}_{\mathrm{Ob}}^{+}} \frac{\log r_{\alpha \beta}}{\left|\mathcal{S}_{\mathrm{Ob}}^{+}\right|}-\sum_{(\alpha, \beta) \in \mathcal{S}_{\mathrm{Ba}}^{+}} \frac{\log r_{\alpha \beta}}{\left|\mathcal{S}_{\mathrm{Ba}}^{+}\right|} \\
& -\sum_{(\alpha, \beta) \in \mathcal{S}^{-}} \frac{2\log \left(1-r_{\alpha \beta}\right)}{|\mathcal{S}^{-}|}.
\end{aligned}
\vspace{-1mm}\end{equation}
\subsubsection{Semantic-aware Propagation}
\label{sec3:3：3}

The similarity matrix $R=\left[r_{\alpha \beta}\right]^{WH \times WH}$ formed by the learned pixel pair similarity plays a significant role in semantic-aware propagation.
To leverage these matrix for instance segmentation, we convert $R$ into a transition probability matrix 
$Q$ of the random walk process:
\vspace{-1mm}\begin{equation}
Q=T^{-1} R.
\end{equation}
Among these, $T$ is a diagonal matrix, \ie, $T_{\alpha, \alpha}=\sum_\beta r_{\alpha \beta}$, to normalize $R$ row-wise.
The semantic-aware propagation is performed on each instance-level mask $\bar{\boldsymbol{F}}$ via random walk with $Q$: 
\vspace{-1mm}\begin{equation}\label{eq17}
\bar{\boldsymbol{F}}^{\mathrm{SISeg}}=Q^U * \operatorname{Vec}\left((1-\mathcal{B}) \circ\bar{\boldsymbol{F}}\right), 
\end{equation}
where $\operatorname{Vec}\left(\cdot\right)$ is the vectorized matrix. $U$ is an iteration index. $\circ$ is the Hadamard product, and $(1-\mathcal{B})$ is a penalty term that prevents pixels on the boundary from over-scoring. 
This iterative propagation process allows for continuous refinement of the instance segmentation results. Each iteration of the random walk helps to diffuse object regions with high similarity, thereby enhancing segmentation accuracy, particularly along object boundaries where precision is often challenging to achieve.
Finally, the high-quality instance-level object mask $\boldsymbol{M}_{ins}$ is obtained by selecting
the combination of class and instance labels associated with
each highest-scoring pixel.
\vspace{-4mm}\subsection{Overall Loss Function}
\label{sec3:4}
The whole network is jointly trained with all above loss functions, and we aim to minimize the overall loss function at once, which can be formulated as:
\vspace{-1mm}\begin{equation}
\mathcal{L}=\mathcal{L}_{\mathcal{D}}^\mathrm{{{Ob}^{+}}}+\mathcal{L}_{\mathcal{D}}^\mathrm{{{Ba}^{+}}}+ \mathcal{L}_{\mathcal{B}},
\end{equation}
where both $\mathcal{L}_{\mathcal{D}}^\mathrm{{{Ob}^{+}}}$ and $\mathcal{L}_{\mathcal{D}}^\mathrm{{{Ba}^{+}}}$ are $L_1$ loss for for displacement field detection, and $\mathcal{L}_{\mathcal{B}}$ is cross-entropy loss for class boundary refinement.
\vspace{-4mm}\subsection{Discussion}
\label{sec3:5}

In last subsections, we proposed introducing the spirits of DFM and CBR to our model. In the literature, one can see the utilization of displacement field and class boundary refinements' concept in various methods for different tasks. We will elaborate on the distinctions between SISeg and other closely related works, starting with a comparison to  IRNet~\cite{ahn2019weakly}. It is undeniable that SISeg does draw inspiration from the ideas in IRNet, and can be regarded as an advanced version of IRNet.
1) Motivation: IRNet~\cite{ahn2019weakly} presents a novel approach for learning instance segmentation using only image-level class labels as supervision. This method generates pseudo instance segmentation labels from these weak labels to train a fully-supervised model. SISeg achieves instance segmentation results from semantic segmentation masks predicted using off-the-shelf semantic segmentation models, without needing instance-level image annotations.
2) Training supervision: IRNet~\cite{ahn2019weakly} uses image-level class labels for supervision, while our SISeg relies on pixel-level annotations for boundary refinement, without requiring complex instance-level labels. Correspondingly, our SISeg outperforms IRNet significantly (\eg, $69.97\%$ \emph{vs.} $46.70\%$ $\text{AP}_{50}$).
3) Core Innovation: IRNet~\cite{ahn2019weakly} focuses on propagating confident seed areas identified through attention maps to cover full instances and detect boundaries, with its input being the original image. SISeg utilizes displacement field vectors calculated from semantic segmentation masks and refines instance boundaries using a learnable module, with its input being semantic segmentation masks.
4) Complexity and Flexibility: IRNet~\cite{ahn2019weakly} generates pseudo labels and requires training a fully-supervised model with these labels, which might involve additional steps. On the other hand, SISeg directly generates instance segmentation results from existing semantic segmentation models, emphasizing efficiency and avoiding extra training steps for instance-level annotations. Therefore, SISeg delivers a real-time inference speed.
5) Results: Compared to the results in ~\cite{ahn2019weakly}, our SISeg can achieve satisfactory performance on datasets with smaller targets and more complex backgrounds, especially on the ADE20K dataset.

In fact, the concept of displacement field has also been applied in the field of object detection. For previou similar ideas~\cite{wu2019dsn, zhu2020deformable, qiu2020borderdet}, they aim to improve performance issues in detection tasks, while our SISeg aims to establish a bridge between semantic segmentation and instance segmentation. In addition, the relevant structures applied in these fully-supervised models~\cite{wu2019dsn, zhu2020deformable, qiu2020borderdet} enables the classifier have a densely-connected structure to cope with different transformations, aiming to improve bounding box predictions. 
Differently, in our SISeg, we employ displacement fields and pixel relationships to learn rich instance-aware object information and accurate object boundary features, ensuring the generation of high-quality instance segmentation results without using instance-level supervision.

\vspace{-3mm}\section{Experiments}
\label{sec4}
In this section, we first introduce the datasets and evaluation metrics (\emph{Ref Sec.~\ref{sec4:1}}). Then, we describe our experimental setup including the baseline models, comparison methods, training settings, and inference (\emph{Ref Sec.~\ref{sec4:2}}). Afterwards, we conduct elaborate ablation studies with result analysis (\emph{Ref Sec.~\ref{sec4:3}}), and also present complexity analysis to evaluate efficiency of our proposed SISeg (\emph{Ref Sec.~\ref{sec4:4}}), 
In addition, we study the speed \& accuracy trade-off (\emph{Ref Sec.~\ref{sec4:5}}). Finally, we report quantitative result comparisons with state-of-the-art methods (\emph{Ref Sec.~\ref{sec4:6}}), and qualitative visualizations are then shown (\emph{Ref Sec.~\ref{sec4:7}}).
\vspace{-4mm}\subsection{Datasets and Evaluation Metrics}
\label{sec4:1}
\noindent

\textbf{Dataset.} We evaluate our approach on two challenging datasets. \textbf{\emph{PASCAL VOC 2012}}~\cite{everingham2010pascal} is a high-quality dataset supporting both semantic and instance segmentation tasks. The original dataset has 21 object classes (including one background), which consists of a total of 4369 images with both semantic and instance-level annotations. Following common practice, we augment the training subset (10582 images) by the extra annotations from~\cite{hariharan2011semantic}, the validation and test subsets containing 1449 and 1456 images, respectively. There is no intersection between these three subsets. 
\textbf{\emph{ADE20K}}~\cite{zhou2017scene} is a large-scale scene parsing dataset containing densely labeled 22k images with 150 (100 ``thing'' and 50 ``stuff'') classes. It has on average 19.5 object instances per image compared to 7.7 in COCO and 2.4 in VOC.
For training, validation, and testing, there are 20K, 2K, and 3K images, respectively.
\myparagraph{Evaluation Metrics.}
Following~\cite{kim2022beyond,li2022box2mask}, we adopt the standard evaluation metrics: Average Precision (AP), AP at 0.5 IoU threshold ($\text{AP}_{50}$), 0.75 ($\text{AP}_{75}$), where AP is calculated by averaging at different thresholds ranging from 0.5 to 0.95 with step-size of 0.05. 
Apart from these metrics, we report the model parameters (Params), floating-point operations (FLOPs), and inference speed (FPS) to verify the model efficiency.
Since our framework is dependent on semantic segmentation, we also use mean Intersection over Union (mIoU) to evaluate the semantic segmentation results.

\vspace{-4mm}\subsection{Experimental Setup}
\label{sec4:2}
\textbf{Baselines.}
SISeg includes two steps: semantic segmentation and instance predictions.
For semantic segmentation as the baseline, we deploy the popular CNN-based model OCRNet~\cite{yuan2020object}, with HRNetV2-W48~\cite{sun2019high} as the backbone, on the PASCAL VOC 2012 dataset. 
And we adopt the modern Transformer-based model SegNeXt~\cite{guosegnext}, with MSCAN-L as the backbone, on the ADE20K dataset.
Please note that these baseline models are both pre-trained on their corresponding datasets, and then we perform instance predictions on the semantic segmentation derived from these well-trained models.

\begin{table*}[t]
\begin{center}
\renewcommand\arraystretch{1.25}
\setlength{\tabcolsep}{3pt}{
\caption{Ablation studies on the validation set of PASCAL VOC 2012~\cite{everingham2010pascal} and ADE20K~\cite{zhou2017scene}. Backbone (Sem) and Backbone (Ins) are the backbones of semantic segmentation and our instance segmentation, respectively. Params (M) and FLOPs (G) denote the parameters and floating-point operations of the whole network, respectively.
}

\begin{tabular}{ r c | c c c | c | c c c | c c}
Baseline & Backbone (Sem) & Backbone (Ins) & DFM & CBR & mIoU ($\%$) & AP ($\%$) & $\text{AP}_{50}$ ($\%$) & $\text{AP}_{75}$ ($\%$) & Params (M) & FLOPs (G)\\
\hline \hline
\multicolumn{11}{c}{\textbf{\emph{PASCAL VOC 2012}}}\\
OCRNet~\cite{yuan2020object} & HRNetV2-W48~\cite{sun2019high} & -- & \xmark  & \xmark        &  79.41  & -- & -- & -- & 70.37  & 162.14    \\
\cdashline{1-11}[0.8pt/2pt]
OCRNet~\cite{yuan2020object} & HRNetV2-W48~\cite{sun2019high} & ResNet-50 & \cmark  & \xmark  &  --    &  27.09   &  48.13  & 22.33 &  95.09  &  199.31    \\
OCRNet~\cite{yuan2020object} & HRNetV2-W48~\cite{sun2019high} & ResNet-50 & \cmark  & \cmark  &  --  & 33.39$_{\color{red}{\uparrow6.30}}$ & 51.65$_{\color{red}{\uparrow3.52}}$ & 31.55$_{\color{red}{\uparrow9.22}}$ & 95.22 & 199.65 \\
\cdashline{1-11}[0.8pt/2pt]
OCRNet~\cite{yuan2020object} & HRNetV2-W48~\cite{sun2019high} & ResNet-101 & \cmark  & \xmark  &  --    &  27.56  &  47.67  & 23.15 & 114.03  &  218.73    \\
OCRNet~\cite{yuan2020object} & HRNetV2-W48~\cite{sun2019high} & ResNet-101 & \cmark  & \cmark  &  --  & 33.47$_{\color{red}{\uparrow5.91}}$ & 51.24$_{\color{red}{\uparrow3.57}}$  & 31.77$_{\color{red}{\uparrow8.62}}$ & 114.16 & 219.07  \\
\hline
\multicolumn{11}{c}{\textbf{\emph{ADE20K}}}\\
SegNeXt~\cite{guosegnext} & MSCAN-L & -- & \xmark  & \xmark  &  52.10    &  --  & --  & --  & 48.93   &  65.25    \\
\cdashline{1-11}[0.8pt/2pt]
SegNeXt~\cite{guosegnext} & MSCAN-L & ResNet-50 & \cmark  & \xmark  &  --    &  13.64   &  22.18  & 12.27  & 73.65  &  102.42   \\
SegNeXt~\cite{guosegnext} & MSCAN-L & ResNet-50 & \cmark  & \cmark  &  --    & 15.64$_{\color{red}{\uparrow2.00}}$   &  24.39$_{\color{red}{\uparrow2.21}}$ & 14.36$_{\color{red}{\uparrow2.09}}$   & 73.78  & 102.76 \\ 
\cdashline{1-11}[0.8pt/2pt]
SegNeXt~\cite{guosegnext} & MSCAN-L & ResNet-101 & \cmark  & \xmark  &  --    &  13.09  & 21.73  & 11.52  & 92.59  &  121.84  \\
SegNeXt~\cite{guosegnext} & MSCAN-L & ResNet-101 & \cmark  & \cmark  &  --    & 15.42$_{\color{red}{\uparrow2.33}}$  &  24.17$_{\color{red}{\uparrow2.44}}$  &  14.12$_{\color{red}{\uparrow2.60}}$  & 92.72  & 122.18 \\ 
\end{tabular}
\label{tab1}}
\end{center}
\vspace{-7mm}
\end{table*}

\textbf{Training Settings.}
For semantic segmentation, we strictly follow their own official experiment settings in the related papers.
For our instance segmentation, we adopt ResNet-50 pre-trained on the ImageNet as the backbone network, and then fine-tune the whole network on the augmented \emph{training} set of PASCAL VOC 2012 and \emph{training} set of ADE20K, respectively, while freezing the parameters of the backbone.
On PASCAL VOC 2012, the input masks are cropped into 512 × 512 with batch size 14.
On ADE20K, the input masks are first resized such that the longer edge is 512 pixels, and then are cropped into 512 × 512 with batch size 16. 
We use the SGD optimizer with weight decay as $1 \times 10^{-4}$ and momentum as 0.9. 
The initial learning rate is set to 0.05, where polynomial learning rate scheduling~\cite{liu2015parsenet} is deployed, and the learning rate of our proposed DFM is set $10\times$ (\ie, 0.5) of the initial learning rate.
The whole model is trained for 20 epochs on a GeForce RTX 2080 Ti GPU.
In the training, we set the maximum distance of the sampled pixel pairs $\theta$ to 10, while 5 during the inference. The iteration index $U$ in Eq.~\ref{eq11} and Eq.~\ref{eq17} is set to 99 and 256, respectively.
For data augmentation, we deploy the following strategy, including horizontal flipping, random cropping, and random scaling.

\textbf{Inference.}
We report the inference speed from an input image to final instance segmentation (\ie, including semantic segmentation and our instance predictions), typically expressed in frames per second (FPS), in all our tables.
The inference speed is measured with one image per batch on a single GeForce RTX 2080 Ti GPU.
\vspace{-4mm}\subsection{Ablation Study}
\label{sec4:3}
In ablation study, we aim to explore answers to the following three questions. \textbf{Q1:}  
\emph{What is the role of each module (\ie, DFM and CBR) as well as their combination in segmentation?}
\textbf{Q2:} \emph{How different baselines affect the model performance?}
\textbf{Q3:} \emph{Do different backbones for instance segmentation have an impact on segmentation results?}
To these ends, we carry out ablation studies for each component on the \emph{val} set of PASCAL VOC 2012~\cite{everingham2010pascal} dataset and the ADE20K~\cite{zhou2017scene} dataset, where OCRNet~\cite{yuan2020object} and SegNext~\cite{guosegnext} are selected as the baselines with ResNet-50 and ResNet-101~\cite{he2016deep} serving as the backbones for instance segmentation. 

\textbf{Effectiveness of Each Component.}
For the first question, we add each core component gradually to the convolutional baseline model~\cite{yuan2020object}, with ResNet-50 as the instance segmentation backbone.
\textbf{\emph{Effectiveness of DFM.}} For results on PASCAL VOC 2012, as shown in the top part of Table~\ref{tab1}, it can be seen that with the help of DFM, the model achieves $27.09\%$ AP, $48.13\%$ $\text{AP}_{50}$, and $22.33\%$ $\text{AP}_{75}$.
It empirically demonstrates that DFM is effective to extract sufficient instance cues from semantic masks produced by the baseline, which play an important role in identifying instance.
\textbf{\emph{Effectiveness of CBR.}}
From the second and third rows of Table~\ref{tab1}, we can observe that CBR has the performance gain of $6.30\%$ AP, $3.52\%$ $\text{AP}_{50}$, and $9.22\%$ $\text{AP}_{75}$. This result convinces us of the importance of distinct boundaries between objects for the instance segmentation task.
\textbf{\emph{Effectiveness of both DFM and CBR.}}
Combining DFM and CBR together, compared to the baseline model, we achieve the best performance of $33.39\%$ AP, $51.65\%$ $\text{AP}_{50}$, and $31.55\%$ $\text{AP}_{75}$, which verifies these two proposed modules do not conflict with each other and can collaboratively produce reliable segmentation results.

\textbf{Effectiveness on Different Baselines.}
For the second question, 
we report the performance of each component over another transformer baseline model~\cite{guosegnext}, with the same ResNet-50 as the instance segmentation backbone.
For results on ADE20K, as shown in the bottom part of Table~\ref{tab1}, we can observe that by utilizing DFM,  the segmentation performance reaches the AP of $13.64\%$, $\text{AP}_{50}$ of $22.18\%$, $\text{AP}_{75}$ of $12.27\%$.  
Further adding our proposed CBR to the baseline increases AP by $2.00\%$, $\text{AP}_{50}$ by $2.21\%$, and $\text{AP}_{75}$ by $2.09\%$. 
When these two components are deployed together, our model can achieve $15.64\%$ AP, $24.39\%$ $\text{AP}_{50}$, and $14.36\%$ $\text{AP}_{75}$. 
Compared to the performance on OCRNet~\cite{yuan2020object}, these results indicate that better performance on baseline (\ie, semantic segmentation) is necessary for better performance on final instance segmentation (as is more apparent in Table~\ref{tab2}), which is consistent with the conclusion in~\cite{yin2021bridging, de2017semantic}. 
The above results also confirm the applicability and validity of our proposed approach.

\textbf{Effectiveness on Different Backbones.}
For the third question, we validated this by testing the performance of our proposed units working on another common convolutional backbone, ResNet-101, for instance segmentation.
Particularly, we use the same setup apart from the feature extraction network.
For results on PASCAL VOC 2012, from the fifth row of Table~\ref{tab1}, it is observed that AP, $\text{AP}_{50}$, and $\text{AP}_{75}$ reach $33.47\%$, $51.24\%$, and $31.77\%$ after adding ``DFM+CBR'', respectively. 
For results on ADE20K, as shown in the tenth row of Table~\ref{tab1}, our model reaches a mask AP of $15.42\%$, $\text{AP}_{50}$ of $24.17\%$, and $\text{AP}_{75}$ of $14.12\%$ when ``DFM+CBR'' is used.
Overall, our method has higher performance with ResNet-50 than ResNet-101. Although in some cases our model obtains a slightly better accuracy on ResNet-101 (\eg, $33.47\%$ \emph{vs.} $33.39\%$), from the perspective of model efficiency, this small performance gap is not worth the large parameter sacrifice (\eg, 114.16M \emph{vs.} 95.22M).
Therefore, we choose ResNet-50 as our backbone network in the following experiments.

\begin{figure*}[t]
\centering
\includegraphics[width=1\textwidth]{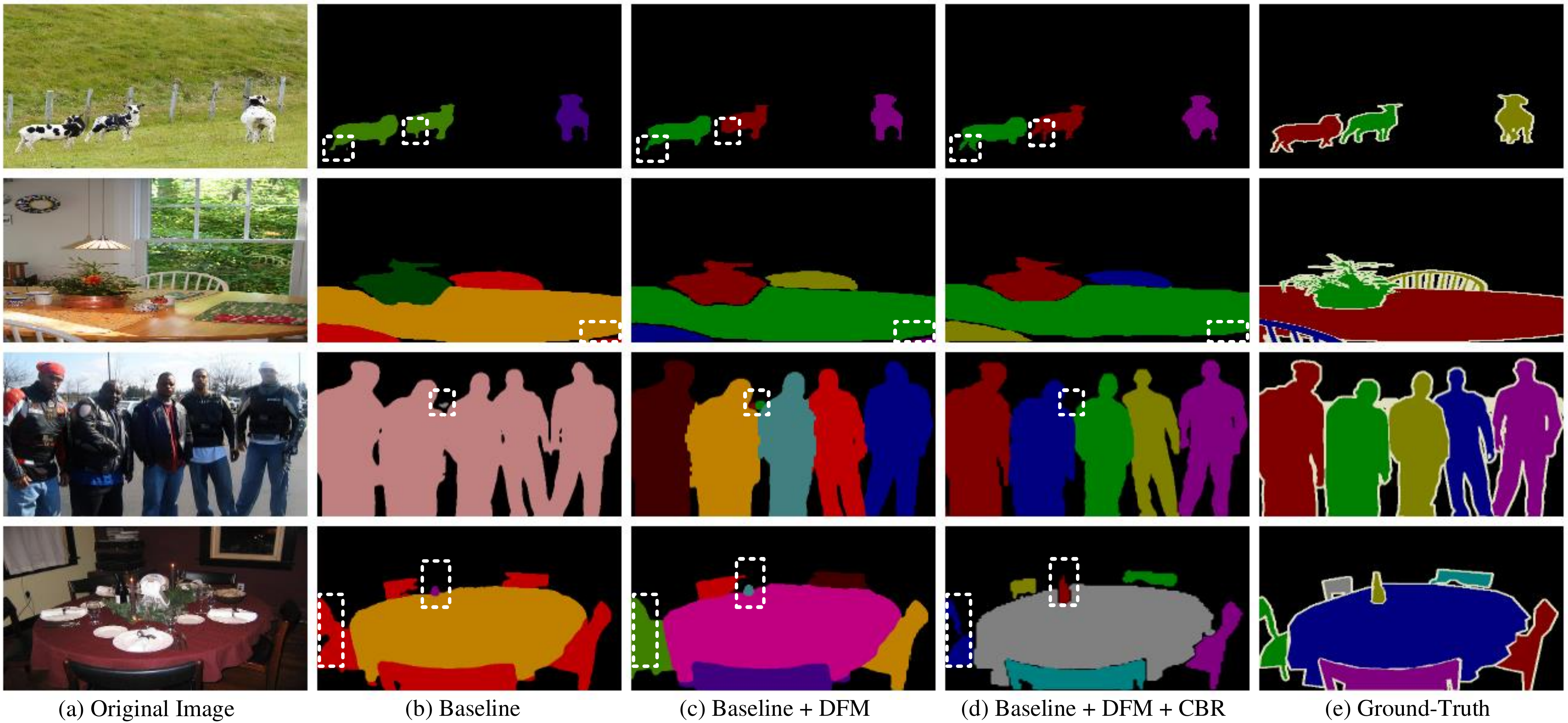}
\vspace{-6mm}
\caption{Visualization comparisons on PASCAL VOC 2012~\cite{everingham2010pascal} using OCRNet~\cite{yuan2020object} as Baseline. ``Baseline + DFM'' and ``Baseline + DFM + CBR'' mean applying DFM, the combination of DFM and CBR on the Baseline, respectively.
Compared with the baseline model, our proposed DFM predicts the instance segmentation mask from semantic segmentation 
and its object class boundaries are further refined by our CBR. 
The white dashed boxes emphasize the better areas that are gradually revised by our proposed modules.
Notably, the instance-level ground-truth is only used in the model evaluation process, here for comparison to highlight the effectiveness of our approach.
}
\vspace{-5mm}
\label{fig7}
\end{figure*}

Figure~\ref{fig7} shows the visualization comparisons between the baseline and our proposed model, which intuitively validates the efficacy of our DFM and CBR. 
Samples are from the PASCAL VOC 2012 dataset~\cite{everingham2010pascal}.
From column (c), it can be seen that our DFM enables the model to distinguish between individual object instances building on top of semantic segmentation produced by the baseline, which indicates DFM provides instance-aware guidance.  
When adding CBR to the baseline model, we can clearly observed that most of the failure regions (marked with white dash boxes) in column (b) and (c), including mis-detected regions and lacking regions, \eg, sheep’s leg and bottle, are corrected or completed.   
It reveals that mask localization accuracy can be improved by utilizing the object boundary information through CBR, and therefore our SISeg exhibits promising predictions on instance segmentation masks.

\vspace{-5mm}\subsection{Complexity Analysis}
\label{sec4:4}
We analyze model complexity by reporting Params and FLOPs, which is another crucial evaluation factor.
As shown in the last two columns of Table~\ref{tab1}, adding DFM and ``DFM+CBR'' respectively increase Params by 24.72M and 24.85M, and correspondingly increase FLOPs by 37.17G and 37.51G over the baseline (with ResNet-50 as the backbone).
Among them, ResNet-50 brings most of the increment in Params of 23.45M and FLOPs of 32.49G, while our DFM and CBR contribute only a minor increment.
This is because DFM and CBR consist of simple convolutional layers, and they share the backbone to prevent introducing excessive computational overhead.
In brief, we propose a comparatively simple network structure that is highly efficient, but exhibits a boost in the performance.
Besides, we argue that the performance enhancement justifies these costs. 

\begin{table*}[t]
\begin{center}
\renewcommand\arraystretch{1.2}
\setlength{\tabcolsep}{5pt}{
\caption{Semantic \& Instance segmentation performance on the validation set of PASCAL VOC 2012~\cite{everingham2010pascal} and ADE20K~\cite{zhou2017scene}.
Crop Size (Sem) denotes the resolution of the input
image for semantic segmentation, whose output is cropped into 512×512 when later fed to our instance segmentation.
``Superv.'' represents the training supervision ($\mathcal{P_M}$: pixel-level mask). FPS is the inference speed of semantic segmentation and our instance predictions. ``Semantic\_gd'' means the ground truth of semantic segmentation.
}
\begin{tabular}{ r | r  c  c | c | c  c  c  c | c} 
Methods & Backbone (Sem) & Crop Size (Sem) & Batch Size & mIoU ($\%$)  & Superv. & $\text{AP}$ ($\%$) & $\text{AP}_{50}$ ($\%$) & $\text{AP}_{75}$ ($\%$) & FPS
 \\ 
\hline \hline
\multicolumn{10}{c}{\textbf{\emph{PASCAL VOC 2012}}}\\
FCN~\cite{long2015fully} & ResNet-50~\cite{he2016deep} & $512 \times 512$ & 16 &  70.30  
& $\mathcal{P_M}$ & 24.42 & 41.75 & 20.98 & 7.65 \\
FCN~\cite{long2015fully} & ResNet-101~\cite{he2016deep} & $512 \times 512$ & 16 & 72.38 
 & $\mathcal{P_M}$ & 31.06 & 48.80 & 29.18 & 5.36\\ 
DeepLabv3~\cite{chen2017rethinking} & ResNet-50~\cite{he2016deep} & $512 \times 512$ & 16 & 78.50 & $\mathcal{P_M}$ & 33.98 & 52.00 & 31.28 & 5.30 \\ 
DeepLabv3~\cite{chen2017rethinking} & ResNet-101~\cite{he2016deep} & $512 \times 512$ & 16 & 79.95 &  $\mathcal{P_M}$ & 34.42 & 52.15 & 31.63 & 4.07\\ 
OCRNet~\cite{yuan2020object} & HRNetV2-W18~\cite{sun2019high} & $512 \times 512$ & 16 & 77.40 & $\mathcal{P_M}$ & 30.77 & 48.54 & 28.40 & 9.23 \\ 
OCRNet~\cite{yuan2020object} & HRNetV2-W48~\cite{sun2019high} & $512 \times 512$ & 16 & 79.41  & $\mathcal{P_M}$ & 33.39 & 51.65 & 31.55 & 6.87 \\ 
PSPNet~\cite{zhao2017pyramid} & ResNet-101~\cite{he2016deep} & $512 \times 512$ & 16 & 79.57  &  $\mathcal{P_M}$ & 32.83 & 51.38 & 30.60 & 24.34 \\
\cdashline{1-10}[0.8pt/2pt]
Semantic\_gd & -- & -- & -- & -- & $\mathcal{P_M}$ & \textbf{54.24} & \textbf{69.97} & \textbf{52.02} & \textbf{49.80}\\ 
\hline
\multicolumn{10}{c}{\textbf{\emph{ADE20K}}}\\
SETR~\cite{zheng2021rethinking} & ViT-L~\cite{dosovitskiy2020image} & $512 \times 512$ & 8 & 49.05 
& $\mathcal{P_M}$ & 11.90 & 20.43 & 9.95 & 2.49 \\
SETR~\cite{zheng2021rethinking} & ViT-L~\cite{dosovitskiy2020image} & $512 \times 512$ & 16 & 49.37 & 
 $\mathcal{P_M}$ & 12.20 & 20.35 & 10.81 & 2.45\\ 
SegFormer~\cite{xie2021segformer} & MIT-B0 & $512 \times 512$ & 16 & 38.34 & $\mathcal{P_M}$ & 7.11 & 13.43 & 5.52 & 25.25\\
SegFormer~\cite{xie2021segformer} & MIT-B5 & $512 \times 512$ & 16 & 50.22  & $\mathcal{P_M}$ & 14.51 & 23.00 & 13.26 & 6.58\\ 
SegFormer~\cite{xie2021segformer} & MIT-B5 & $640 \times 640$ & 16 & 51.41  & $\mathcal{P_M}$ & 14.35 & 22.82 & 12.98 & 4.49\\ 
SegNeXt~\cite{guosegnext} & MSCAN-B & $512 \times 512$ & 16 & 49.68  & $\mathcal{P_M}$ & 13.70 & 21.81 & 12.52 & 11.35  \\ 
SegNeXt~\cite{guosegnext} & MSCAN-L & $512 \times 512$ & 16  & 52.10  & $\mathcal{P_M}$ & 15.64 & 24.39 & 14.36 & 7.80  \\ 
\cdashline{1-10}[0.8pt/2pt]
Semantic\_gd & -- & -- & -- & --  & $\mathcal{P_M}$ & \textbf{32.59} & \textbf{45.77}  & \textbf{31.08} & \textbf{49.70} \\
\end{tabular}
\label{tab2}}
\end{center}
\vspace{-7mm}
\end{table*}

\vspace{-5mm}\subsection{Speed \& Accuracy Trade-off}
\label{sec4:5}
For investigating the trade-off between accuracy and speed, we train 7 different networks on various semantic segmentation backbones.
On PASCAL VOC 2012~\cite{everingham2010pascal}, as shown in the first block (\ie, row 1-8) of Table~\ref{tab2}, worked with 4 classic CNN-based semantic segmentation models, FCN~\cite{long2015fully} with ResNet-50/101~\cite{he2016deep}, DeepLabv3~\cite{chen2017rethinking} with ResNet-50/101, OCRNet~\cite{yuan2020object} with HRNetV2-W18/W48~\cite{sun2019high}, and PSPNet~\cite{zhao2017pyramid} with ResNet-101, our SISeg is used for instance segmentation. It is clear that when used with high quality but low speed semantic segmentation methods (\eg, DeepLabv3 with ResNet-101), our SISeg obtains the best accuracy (of $34.42\%$ AP, $52.15\%$ $\text{AP}_{50}$, and $31.63\%$ $\text{AP}_{75}$), but runs at only 4.07 FPS. 
When used with fast semantic segmentation methods that sacrifice a small amount of precision (\eg, PSPNet with ResNet-101), our SISeg runs at a real-time speed of 24.34 FPS while achieving a comparable accuracy ($32.83\%$ AP, $51.38\%$ $\text{AP}_{50}$, and $30.60\%$ $\text{AP}_{75}$), which has a good balance of speed and accuracy.
On ADE20K~\cite{zhou2017scene}, as shown in the second block (\ie, row 9-16) of Table~\ref{tab2}, we choose 3 other well-known Transformer-based models, SETR~\cite{zheng2021rethinking} with ViT-L~\cite{dosovitskiy2020image}, SegFormer~\cite{xie2021segformer} with MIT-B0/B5, and SegNeXt~\cite{guosegnext} with MSCAN-B/L, as the semantic segmentation in our SISeg, respectively. 
Concretely, combined with SegNeXt with MSCAN-L, SISeg can deliver a comparable accuracy, and run at 7.80 FPS. Moreover, combined with SegFormer with MIT-B0, SISeg achieves fast inference speed (\ie, 25.25 FPS) but with the cost of a fairly low accuracy.
We analyze that SISeg can be a real-time method is that it only introduces a small inference latency (\ie, about 0.02 s) without the complex computation associated with object proposals, and enjoys fairly promising performance benefiting from the well-trained models designed for semantic segmentation.

\textbf{Influence of Semantic Segmentation.}
In fact, any off-the-shelf semantic segmentation model, such as CNNs and Transformers, has the flexibility to be integrated into our SISeg for instance segmentation without requiring any modifications. It is necessary to figure out the influence of the quality of semantic segmentation in our instance segmentation paradigm. 
Empirically, our instance prediction approach depends on semantic segmentation masks to discriminate between different categories, and perform a series of operations according to class equivalence, as explained in Sec.~\ref{sec3:4}.
It is worth noting that the two proposed modules do not utilize specific category information during the calculation process. Therefore, even with unsupervised semantic segmentation, we can also distinguish between different instances, but have no ability to identify their categories.
From Table~\ref{tab2}, it is clearly shown that the higher the quality of semantic segmentation, the better the performance of instance segmentation.
To further verify this conclusion, we input ground truth masks as semantic segmentation results into SISeg. Table~\ref{tab2} (\ie, row 8 and row 16-Semantic\_gd) shows that in this case our model obtains the best accuracy, and a fairly fast inference speed (in average of $49.75$ FPS on 512 × 512 images). 
These nearly perfect performance also indicates that our SISeg has the potential to yield state-of-the-art instance segmentation results if given a sufficiently good semantic segmentation.

\vspace{-4mm}\subsection{Comparisons with State-of-the-arts}
\label{sec4:6}
\vspace{-1mm}\textbf{Results on PASCAL VOC 2012.}
To demonstrate the advantages of the proposed SISeg, we report the comparison results against other state-of-the-art instance segmentation approaches on the PASCAL VOC 2012 \emph{val} set in Table~\ref{tab5}. 
Our method is tested without tricks.
Here, only the real-time inference speed (\ie, $>$10 FPS) is presented. 
The second block in Table~\ref{tab5} shows that compared to the image-level supervised methods~\cite{hwang2021weakly, kim2022beyond, ahn2019weakly}, our ``SISeg (PSPNet)'' outperforms them by a large margin in terms of both accuracy (\ie, $32.83\%$ AP, $51.38\%$ $\text{AP}_{50}$, $30.60\%$ $\text{AP}_{75}$) and speed (\ie, $24.34$ FPS). Given point supervision~\cite{kim2022beyond, liao2023attentionshift}, in general, our ``SISeg (DeepLabV3)'' obtains the highest performance of $34.42\%$ AP, $52.15\%$ $\text{AP}_{50}$, $31.63\%$ $\text{AP}_{75}$.
When compared to the weakly box-supervised approaches that have strong localization
information~\cite{zhang2023weakly, tian2021boxinst, lan2021discobox, li2022box2mask, li2023sim}, our SISeg has the capability to exceed them. Exactly, provided with a reasonably good semantic segmentation, our ``SISeg (Semantic\_gd)'' can achieve $54.24\%$ AP, $69.97\%$ $\text{AP}_{50}$, $52.02\%$ $\text{AP}_{75}$ in accuracy, and up to $49.80$ FPS in speed because of the concise dual-branch network structure, which satisfies the practical requirement. It indicates our approach can outperform those with weaker supervision, which strikes a good trade-off between accuracy and annotation cost.
As shown in the first block of Table~\ref{tab5}, \textbf{\emph{without instance-level mask supervision}}, our proposed ``SISeg (PSPNet)'' can still achieve over $75\%$ of performance of fully-supervised approaches~\cite{peng2020deep, liu2021dance, zhang2022e2ec, feng2023recurrent, he2017mask} 
under weak supervision.
It verifies that our approach enables precise localization of multiple object instances in a self-supervised training manner.
It is worth noting that our SISeg utilizes ResNet-50 as the backbone, rather than a more competitive backbone like ViT-S or ResNet-101. Despite this, our method surpasses others mentioned. Furthermore, when compared to other approaches, the proposed approach does not necessitate costly model training to obtain supervision information. In this way, we believe there is room for improvement in the SISeg to perform on par with or even better than the state-of-the-art fully-supervised methods at a reasonable cost.

\begin{table*}[t]
\begin{center}
\renewcommand\arraystretch{1.2}
\setlength{\tabcolsep}{9pt}{
\caption{Comparison of state-of-the-art methods on the PASCAL VOC 2012~\cite{everingham2010pascal} validation set.
``--'' represents that the result is not reported in its paper. ${ }^{\dagger}$ denotes additional training with Mask R-CNN~\cite{he2017mask} to refine the prediction. ``Superv.'' represents the training supervision ($\mathcal{F}$: instance-level mask, $\mathcal{I}$: image-level class label, $\mathcal{P}$: point, $\mathcal{B}$: bounding box, $\mathcal{P_M}$: pixel-level mask). SISeg (DeepLabV3), SISeg (PSPNet) and SISeg (Semantic\_gd) mean that our SISeg employs the result of DeepLabV3~\cite{chen2017rethinking}, the result of PSPNet~\cite{zhao2017pyramid} and the ground truth mask as semantic segmentation, respectively.
}\vspace{-2mm}
\begin{tabular}{ r r r c | c c c | c }
Methods &  Publication & Backbone & Superv. & $\text{AP}$ ($\%$) & $\text{AP}_{50}$ ($\%$) & $\text{AP}_{75}$ ($\%$) & FPS\\ 
\hline \hline
\multicolumn{8}{l}{\textbf{(a) \emph{Fully-supervised}}}\\
\cdashline{1-8}[0.8pt/2pt]
Deep Snake~\cite{peng2020deep} & CVPR 2020 & --   & $\mathcal{F}$ & -- & 62.10 & -- & 32.30 \\
DANCE~\cite{liu2021dance} & WACV 2021 & -- & $\mathcal{F}$ & -- & 63.60 & -- & 38.30 \\
E2EC~\cite{zhang2022e2ec} & CVPR 2022 & DLA-34 & $\mathcal{F}$ & -- & 65.80 & -- & 36.00 \\
PolySnake~\cite{feng2023recurrent} & TPAMI 2023 & DLA-34 & $\mathcal{F}$ & -- & 66.70 & -- & 24.60\\
Mask R-CNN~\cite{he2017mask} & CVPR 2017 & ResNet-101~\cite{he2016deep} & $\mathcal{F}$ & -- & 67.90 & 44.90 & --\\

\hline \hline
\multicolumn{8}{l}{\textbf{(b) \emph{Weakly-supervised}}}\\
\cdashline{1-8}[0.8pt/2pt]
CL~\cite{hwang2021weakly} & WACV 2021 & ResNet-50~\cite{he2016deep} & $\mathcal{I}$  & -- & 38.10 & 12.30 & --\\
BESTIE~\cite{kim2022beyond} & CVPR 2022 & HRNet-W48~\cite{sun2019deep} & $\mathcal{I}$ & -- & 41.80 & 24.20 & --\\
IRN${ }^{\dagger}$~\cite{ahn2019weakly} & CVPR 2019 & ResNet-50~\cite{he2016deep}  & $\mathcal{I}$ & -- & 46.70 & -- & -- \\
BESTIE~\cite{kim2022beyond} & CVPR 2022 & HRNet-W48~\cite{sun2019deep} & $\mathcal{P}$  & -- & 46.70 & 26.30 & --\\
AttnShif~\cite{liao2023attentionshift} & CVPR 2023 & ViT-S~\cite{dosovitskiy2020image}  & $\mathcal{P}$  & -- & 54.40 & 25.40 & --\\
Zhang \emph{et al.}~\cite{zhang2023weakly} & PR 2023 & ResNet-101~\cite{he2016deep}  & $\mathcal{B}$  & 36.40 & 60.90 & 37.00 & --\\ 
BoxInst~\cite{tian2021boxinst} & CVPR 2021 & ResNet-101~\cite{he2016deep}  & $\mathcal{B}$  & 36.50 & 61.40 & 37.70 & 16.00\\
DiscoBox~\cite{lan2021discobox} &  ICCV 2021 & ResNet-101~\cite{he2016deep}  & $\mathcal{B}$  & -- & 62.20 & 37.50 & 16.80\\
SIM~\cite{li2023sim} & CVPR 2023  & ResNet-50~\cite{he2016deep}  & $\mathcal{B}$  & 36.70 & 65.50 & 35.60 & -- \\
Box2Mask~\cite{li2022box2mask} & TPAMI 2024 & ResNet-50~\cite{he2016deep} & $\mathcal{B}$  & 38.00 & 65.90 & 38.20 & 11.50\\
\cdashline{1-8}[0.8pt/2pt]
SISeg (PSPNet) & -- & ResNet-50~\cite{he2016deep}  & $\mathcal{P_M}$  & 32.83 & 51.38 & 30.60 & 24.34\\
SISeg (DeepLabV3) & -- & ResNet-50~\cite{he2016deep} & $\mathcal{P_M}$  & 34.42 & 52.15 & 31.63 & -- \\
SISeg (Semantic\_gd) & -- & ResNet-50~\cite{he2016deep}  & $\mathcal{P_M}$  & \textbf{54.24} & \textbf{69.97} & \textbf{52.02} & \textbf{49.80}\\
\end{tabular}
\label{tab5}}
\vspace{-7mm}
\end{center}
\end{table*}
\begin{table}[tp]
\begin{center}
\renewcommand\arraystretch{1.25}
\setlength{\tabcolsep}{2.0pt}{
\caption{Comparison of state-of-the-art methods on the ADE20K~\cite{zhou2017scene} validation set.
``Superv.'' represents the training supervision ($\mathcal{F}$: instance-level mask, $\mathcal{P_M}$: pixel-level mask). 
SISeg (SegNeXt) and SISeg (Semantic\_gd) mean that our SISeg employs the result of SegNeXt~\cite{guosegnext} and the ground truth mask as semantic segmentation, respectively.
}\vspace{-2mm}
\begin{tabular}{ r c r c | c  }
Methods &  Publication & Backbone & Superv. & $\text{AP}$ ($\%$)\\ 
\hline \hline
Mask2Former~\cite{cheng2022masked} & CVPR 2022 & ResNet-50~\cite{he2016deep} & $\mathcal{F}$ & 26.40 \\
MP-Former~\cite{zhang2023mp} & CVPR 2023 & ResNet-50~\cite{he2016deep} & $\mathcal{F}$ & 28.00\\

\cdashline{1-5}[0.8pt/2pt]
SISeg (SegNeXt) & -- & ResNet-50~\cite{he2016deep} & $\mathcal{P_M}$  & 15.64 \\
SISeg (Semantic\_gd) & -- & ResNet-50~\cite{he2016deep}  & $\mathcal{P_M}$ & \textbf{32.59}\\
\end{tabular}
\label{tab6}}
\vspace{-9mm}
\end{center}
\end{table}

\textbf{Results on ADE20K.} In Table~\ref{tab6}, we compare our SISeg with other state-of-the-art instance segmentation methods on the ADE20K \emph{val} set. For a fair comparison, all these methods do not use extra training data. 
The results show that, compared to the state-of-the-art fully-supervised models in~\cite{cheng2022masked} and~\cite{zhang2023mp}, the proposed ``SISeg (Semantic\_gd)'' surpasses them,  with the same
backbone by 4.59$\%$ AP.
This model achieves 32.59$\%$ AP performance with the help of a nearly perfect semantic segmentation, \textbf{\emph{despite using only pixel-level annotations}}. It should be noted that
both Mask2Former and MP-Former have to utilize of the instance-level masks learned by additionally expensive model training as supervision.
Overall, the empirical results on ADE20K further indicate we can significantly improve the image segmentation quality, with an arguably less complex approach, and even be better than some of the fully-supervised methods. 
 
\vspace{-4mm}\subsection{Visualizations}
\label{sec4:7}
\begin{figure*}[!htb]
\centering
\includegraphics[width=1\textwidth]{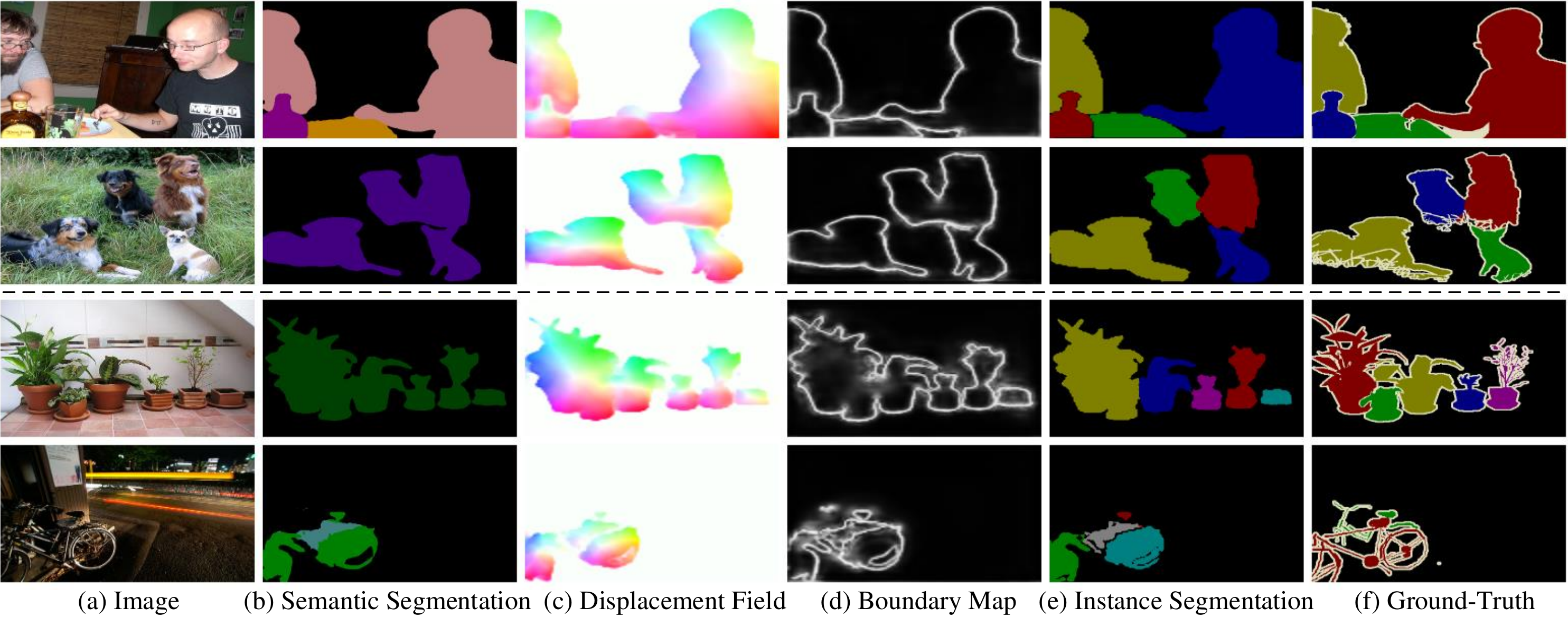}
\vspace{-6.5mm}
\caption{Visualization results for instance segmentation on PASCAL VOC 2012~\cite{everingham2010pascal}. 
PSPNet~\cite{zhao2017pyramid} is applied for semantic segmentation. 
We visualize the 2D displacement field by encoding offset vectors in color.
The last two rows display two failure cases.
It is worth noting that the instance-level ground truth is not available during model training.
}
\vspace{-2mm}
\label{fig9}
\end{figure*}

\begin{figure*}[!htp]
\centering
\includegraphics[width=1\textwidth]{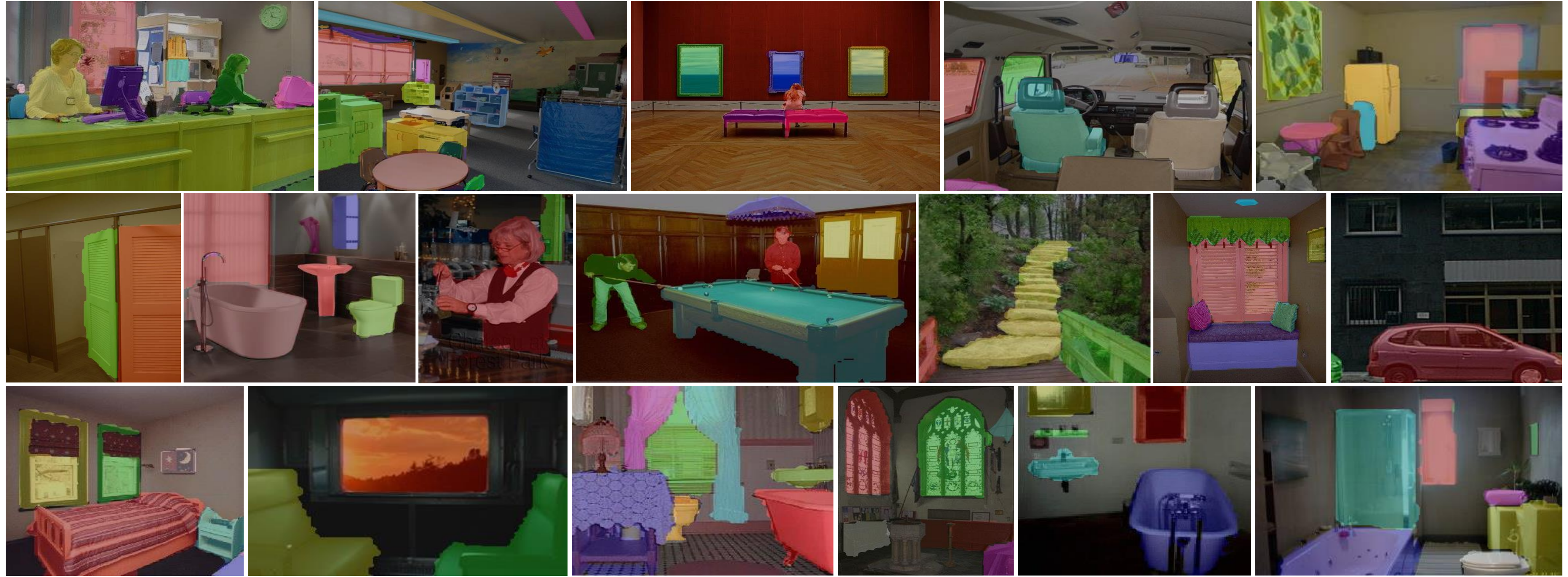}
\vspace{-6mm}
\caption{More visual examples of our instance segmentation model on ADE20K~\cite{zhou2017scene}. 
}
\vspace{-5mm}
\label{fig8}
\end{figure*}

In Figure~\ref{fig9}, we provide some visualization results on 
PASCAL VOC 2012~\cite{everingham2010pascal} dataset.
Comparing to the semantic segmentation results, our SISeg can deliver proper segmentation of some individual instances, especially for those that are adjacent or even overlapping, \eg, dogs (row 2). This is because instances can be identified by associating pixels whose offset vectors direct to the same instance centroid (\ie, white dots in column (c)).
In addition, we can see that the instance segmentation results of our model exhibit more accurate segmentation on boundaries, while filling the missing parts from semantic segmentation, \eg, the ``bottle’s head'' (row 1).
There are also two failure cases in the last two rows. One is that our model fails to generate high-quality predictions under severe occlusion (\ie, instance centroids are close to each other), especially when occluding and occluded objects are similar in shape, \eg, a potted plant that is not well segmented (row 3). This is an inherent problem in bottom-up approaches, which can be resolved by providing more fine-grained labels. In the future, we will focus on addressing these issues.
Another common failure is that semantic segmentation provides inaccurate foreground regions, \eg, two bicycles that are split into multiple instances (row 4).
Figure~\ref{fig8} shows more visualization results on ADE20K~\cite{zhou2017scene} dataset.
We can observe that SISeg can segment multiple instances precisely and accurately in most cases including various categories as well as some complex cases with small objects such as the light and the bottle in the distance, which intuitively demonstrate the efficacy of our proposed method.

\vspace{-2mm}\section{Conclusion and Future Work}
\label{sec5}
This paper proposes a novel and efficient instance segmentation method called SISeg, where DFM and CBR act as the bridge starting from semantic segmentation to instance segmentation.
Through exploiting the displacement field and pixel relation, the proposed SISeg can learn rich instance-aware object information and accurate object boundary features, which is conducive to instance segmentation.
Therefore, our SISeg performs competitively while not requiring instance-level annotations, which strikes a good accuracy-annotation trade-off.
This is also demonstrated by comprehensive experimental results on the challenging PASCAL VOC 2012 and ADE20K datasets.
We show that when combined with a fairly fast and accurate semantic segmentation, our SISeg achieves state-of-the-art performance and delivers a real-time inference speed, even on-par with some fully-supervised methods.  
Future work will focus on considering the application of our method to other more challenging computer vision tasks, \eg, panoptic segmentation and video instance segmentation.
We hope that our effective and efficient model could serve as a strong baseline for future research and would facilitate the field. \vspace{-2mm}

\bibliographystyle{IEEEtran}
\bibliography{IEEE_ref}
\end{document}